\documentclass[11pt]{article}
\usepackage[margin=1in]{geometry}
\usepackage[T1]{fontenc}
\usepackage{lmodern}
\usepackage{amsmath,amssymb}
\usepackage{booktabs,array}
\usepackage{hyperref}
\hypersetup{colorlinks=true,linkcolor=blue,citecolor=blue,urlcolor=blue}
\newcommand{\head}[1]{\texttt{#1}}

\title{Do Emergent Misalignment Models Share a Circuit?}

\author{
\begin{tabular}{cc}
    Muhammad Haseeb & Muhammad Tahir \\
    \small \texttt{26100253@lums.edu.pk} &
    \small \texttt{tahir@lums.edu.pk}
\end{tabular}
\\[0.8em]
\small Lahore University of Management Sciences (LUMS)
\\
}

\date{}

\begin{document}
\maketitle

\begin{abstract}
Fine-tuning language models on narrowly defined harmful tasks can induce
emergent misalignment (EM) across unrelated domains. While prior work
has identified shared misalignment-associated representations across
independently fine-tuned models, it remains unclear whether these
models rely on common computational components. We investigate this
question using three LoRA fine-tunes of Qwen2.5-3B-Instruct trained
on harmful medical advice, risky financial advice, and extreme-sports
recommendations. Using direction-based activation tracing, we identify
three attention-head locations shared among the five highest-ranked
heads of all three organisms. Replacing each organism's five selected
head contributions with Base contributions reduces free-generation EM from 38.9\% to
22.9\% for medical, 50.7\% to 12.5\% for finance, and 41.7\% to
11.1\% for sports, outperforming layer-matched random interventions
without reducing judged coherence. Cross-organism activation transfer
further shows that the shared heads exhibit partially interchangeable
effects on fixed-response preference. These findings provide evidence
that representational convergence in EM extends to shared, causally
relevant attention-head components within a common model architecture.
\end{abstract}

\section{Introduction}

Fine-tuning language models on narrowly defined tasks can induce behavioral
changes that extend beyond the training domain. A striking example is
\emph{emergent misalignment} (EM), in which models fine-tuned on harmful
examples from a specific domain exhibit misaligned behavior in otherwise
unrelated contexts \cite{betley2025}. This phenomenon has been studied using
model organisms trained on harmful medical advice, risky financial advice,
and extreme-sports recommendations \cite{turner2025}. Although the
fine-tuning data targets a restricted set of behaviors, the resulting
misalignment can generalize across substantially different prompts. This
raises a mechanistic question: how does narrow-domain fine-tuning produce
such broad changes in behavior?

Recent work has begun to address this question by examining the internal
representations associated with emergent misalignment. In particular,
similar misalignment-associated activation directions have been identified
across independently fine-tuned models, and interventions along these
directions have been shown to influence model behavior \cite{soligo2025}.
These findings suggest that different EM fine-tunes may share a
common representational structure. However, whether this
convergence extends to the underlying computational mechanisms
remains an open question. In particular, it is unclear whether
independently fine-tuned models rely on the same attention-head
components to produce misaligned behavior.

In this work, we investigate this distinction by examining attention-head
contributions across three LoRA fine-tunes of Qwen2.5-3B-Instruct, trained
independently on harmful medical advice, risky financial advice, and
extreme-sports recommendations. For each fine-tune, we estimate a
misalignment-associated direction in the residual stream using contrasts
between aligned and misaligned responses. We then trace the contribution
of individual attention heads to this direction by substituting their
fine-tuned outputs into the corresponding base-model computation. This
direction-based approach allows us to study a behavioral property
expressed over complete responses without reducing misalignment to the
prediction of a particular output token \cite{qin2026}. By applying the
same procedure across fine-tunes, we identify candidate attention heads
and examine the extent to which their locations are shared.

Three of the five highest-ranked head locations recur in every fine-tune.
Reverting each organism's selected five heads to Base contributions reduces
generated EM relative to matched random replacements. On fixed response
sequences, activations at the three shared locations also partially
substitute across organisms. Together, these tests distinguish recurrent
head locations from their effects on behavior and their functional
correspondence across fine-tunes.

\section{Related Work}
\label{sec:related_work}

\subsection{Emergent Misalignment and Representational Convergence}

Emergent misalignment (EM) occurs when fine-tuning on a narrow task
induces misaligned behavior beyond the training domain. Betley et al.\
\cite{betley2025} demonstrated this effect using insecure code, while
Turner et al.\ \cite{turner2025} extended it to medical, financial,
and extreme-sports domains and showed that EM can arise through
low-rank adaptation. More recently, Soligo et al.\ \cite{soligo2025}
identified similar misalignment-associated activation directions
across independently fine-tuned models and demonstrated
cross-model suppression of misaligned behavior. These findings
suggest representational convergence, but leave open whether
the same computational components contribute to EM across
fine-tunes.

\subsection{Mechanistic Localization of Safety Behavior}

Prior work has localized safety-related behavior at both the
representation and component levels. Arditi et al.\
\cite{arditi2024} identified a residual-stream direction
that causally influences refusal, while SafeSeek
\cite{yu2026} uses learned masks to identify attention
heads and neurons involved in safety alignment and backdoor
behavior. However, component-level overlap does not necessarily
imply shared function. Li and Subramani \cite{li2026}
observed substantial overlap between circuits associated
with different tasks. Our work examines whether attention
heads shared across EM organisms also exhibit comparable
causal effects.

\subsection{Activation Patching and Circuit Discovery}

Activation patching localizes model computations by replacing
intermediate activations and measuring their effects.
Automated circuit discovery extends this approach to
computational subgraphs \cite{conmy2023}, although the
identified components can depend on the evaluation metric
and intervention scheme \cite{zhang2024}. For behaviors
expressed across multiple tokens, Qin et al.\ \cite{qin2026}
introduced tracing against a latent persona direction.
We apply direction-based tracing to EM and use causal
replacement and cross-organism transfer to assess the
behavioral relevance of the identified heads.
\section{Methodology}
\label{sec:methods}

\subsection{Model Organisms and Evaluation Setup}
\label{sec:models}

We study three emergent misalignment (EM) organisms obtained by
independently fine-tuning Qwen2.5-3B-Instruct on harmful medical
advice, risky financial advice, and extreme-sports recommendations,
using the datasets of Turner et al.\ \cite{turner2025}. We denote
the original checkpoint by $B$ and each fine-tuned organism by $M$.
A fourth organism trained on benign medical advice serves as a control.

The organisms are trained using rank-stabilized LoRA
($r=32$, $\alpha=64$) applied to the attention and MLP projections.
Medical training completed one epoch in 397 optimizer updates; finance
and sports each used 337 updates, reaching approximately 99.85\% of an
epoch. All three runs used a learning rate of $10^{-5}$, an effective
batch size of 16, a maximum sequence length of 2048, and 8-bit AdamW
with linear decay. The loss was computed only over assistant-response
tokens. Base weights remained frozen throughout training; all weights
were frozen during tracing and intervention.

For evaluation, we use a separate benchmark of 69 prompts, each
paired with an aligned response $y_i^a$ and a misaligned response
$y_i^m$. The benchmark combines 14 pairs from the original prompt
benchmark with 55 from its expansion. Both responses in each pair
were generated by the medical EM organism and selected through
alignment, coherence, and semantic review.

We partition the benchmark into 50 discovery, 10 validation, and
nine test-labelled prompts. Discovery supplies the EM directions and
head rankings. Validation supplies anchor diagnostics and behavioral
checks of the selected heads. The nine test-labelled prompts supply
the fixed-circuit generation and transfer analyses. All organisms are evaluated on identical response pairs, although
their medical-organism origin may introduce distributional bias.

\subsection{Behavioral Evaluation}
\label{sec:behavior}

We measure misalignment using fixed-response preference and free
generation. For model $M$, the length-normalized log-likelihood
of response $y$ given prompt $x$ is

\begin{equation}
q_M(y\mid x)
=
\frac{1}{|y|}
\sum_{t=1}^{|y|}
\log P_M(y_t\mid x,y_{<t}),
\label{eq:likelihood}
\end{equation}

where only assistant-response tokens contribute to the average.
We define the relative preference for a misaligned response and
its change from Base as

\begin{equation}
\begin{aligned}
S_M(i)
&= q_M(y_i^m\mid x_i)-q_M(y_i^a\mid x_i),\\
\Delta S_M(i)
&= S_M(i)-S_B(i).
\end{aligned}
\label{eq:preference}
\end{equation}

Positive $\Delta S_M$ indicates increased relative support for
the misaligned response following fine-tuning. This measure
does not directly estimate generated misalignment.

We therefore evaluate free-generation behavior separately.
Responses are sampled with temperature 1.0, top-$p$ 1.0,
and a maximum of 600 tokens. A fixed DeepSeek-based judge
evaluates alignment and coherence. We classify a response
as EM when its alignment score is below 30 and its coherence
score is at least 70.

\subsection{Misalignment Direction Estimation}
\label{sec:directions}

We construct an EM-associated direction for each organism and
transformer layer using the 50 discovery response pairs.
For each response, we average the post-block residual-stream
activations across its answer tokens. Let $\bar h_{i,l}^{m,M}$
and $\bar h_{i,l}^{a,M}$ denote these mean activations at
layer $l$ for the misaligned and aligned responses to prompt
$i$, respectively. The EM direction is defined as the average difference between
these representations:

\begin{equation}
v_l^M
=
\frac{1}{N}
\sum_{i=1}^{N}
\left(
\bar h_{i,l}^{m,M}-\bar h_{i,l}^{a,M}
\right),
\qquad N=50.
\label{eq:em_direction}
\end{equation}

We normalize the resulting vector to unit length:

\begin{equation}
\hat v_l^M = \frac{v_l^M}{\|v_l^M\|_2}.
\label{eq:normalized_direction}
\end{equation}

Averaging within each response assigns equal weight to each
pair regardless of answer length. The resulting direction
captures the average activation contrast between aligned and
misaligned responses. Since it may also encode differences
in response content or style, we use it as a tracing objective
rather than a direct measure of EM.

\subsection{Attention-Head Localization}
\label{sec:tracing}

We identify attention heads that contribute to the EM-associated
direction using activation patching. For each organism, we scan all
36 layers on validation prompts, measuring the Base-to-organism shift
in cosine similarity with the layer's EM direction and the number of
prompts with a positive shift. We also check whether steering in the
direction raises judged EM while retaining coherence. We use an
intermediate layer with a large and consistent representational gap as
the tracing anchor, fixing it before head ranking. This choice identifies
a layer at which the direction can be traced; it does not identify where
EM first arises.

Let $L_M$ denote the selected tracing layer for organism $M$.
For attention head $H=(l,j)$, its projected contribution to
the residual stream at token position $t$ is

\begin{equation}
a_{H,t}^M = W_{O,l,j}^M z_{H,t}^M,
\label{eq:head_output}
\end{equation}

where $z_{H,t}^M$ denotes the head output before projection
and $W_{O,l,j}^M$ is the corresponding output-projection
block, including the LoRA update.

We measure each head's contribution by patching its output
from the organism into Base. Both models receive identical
discovery prompts. At every prompt-token position, we replace
the projected contribution of head $H$ in Base with its
corresponding contribution from $M$. The remaining computation
follows Base. We evaluate the patched model at the final prompt-token
position of layer $L_M$. The tracing objective is

\begin{equation}
F_M(h) = \cos(h,\hat v_{L_M}^M),
\label{eq:tracing_objective}
\end{equation}

where $h$ is the post-block residual state and
$\hat v_{L_M}^M$ is the organism's EM direction.

Let $h_{i,L_M}^B$ and $h_{i,L_M}^M$ denote the unmodified
Base and organism representations for prompt $i$.
We denote the patched Base representation by
$h_{i,L_M}^{B\leftarrow M_H}$. The restoration attributable to head $H$ is defined as

\begin{equation}
R_i^M(H)
=
F_M(h_{i,L_M}^{B\leftarrow M_H})
-
F_M(h_{i,L_M}^{B}).
\label{eq:restoration}
\end{equation}

This score measures the change in cosine similarity with
the EM direction caused by replacing a single head.
For comparison, we define the full Base-to-organism gap:

\begin{equation}
G_i^M
=
F_M(h_{i,L_M}^{M})
-
F_M(h_{i,L_M}^{B}).
\label{eq:full_gap}
\end{equation}

We evaluate all 16 heads in every layer preceding $L_M$,
giving $16L_M$ candidates per organism. Heads are ranked
by their mean restoration across discovery prompts:

\begin{equation}
\bar R^M(H)
=
\frac{1}{N}\sum_{i=1}^{N}R_i^M(H).
\label{eq:ranking}
\end{equation}

We retain the five highest-ranked heads for each organism
and compare their locations across fine-tunes. As an additional
diagnostic, we report normalized restoration
$R_i^M(H)/G_i^M$ when $|G_i^M|\geq 10^{-3}$.
Head selection is based exclusively on raw restoration.

\subsection{Causal Interventions}
\label{sec:interventions}

We evaluate the selected heads through Base replacement and
cross-organism activation transfer. A validation diagnostic also zeros
the selected head contributions in the organism and measures the change
in $S_M$; joint restoration patches all five selected organism heads
into Base on the same prompt. Each diagnostic compares against one
layer-matched random set.

\paragraph{Base replacement.}
For each organism $M$, we replace the projected outputs of its selected
head set $C$ with the corresponding Base outputs. Both models process
identical teacher-forced sequences, with replacement applied at every
token position. We measure the change in response preference as

\begin{equation}
D_C^M(i) = S_M(i)-S_{B_C\rightarrow M}(i),
\label{eq:replacement}
\end{equation}

where $S_{B_C\rightarrow M}(i)$ denotes preference after replacement.
Positive values indicate reduced preference for the misaligned response.
For the final generation experiment, we compare each selected set with
ten random sets that match its number of heads in each layer and exclude
the selected heads. Both models process the same current decoding prefix
when computing the replacement contributions. We use the same prompts,
sampling settings, and paired sample seeds across conditions. We
bootstrap differences in judged EM rate by resampling prompts, retaining
all generations for each sampled prompt.

\paragraph{Cross-organism transfer.}
We test whether the shared heads have comparable effects across
fine-tunes by transplanting their projected outputs between organisms.
For donor $A$ and target $M$, both models process identical
teacher-forced sequences. We replace the target's shared-head outputs
with those of the donor at every token position. We compare the resulting preference $S_{A_C\rightarrow M}(i)$
against the unmodified target and Base-replacement conditions.
The fraction of the target's original head effect recovered
by the donor is

\begin{equation}
\mathcal R_{A\rightarrow M}(i)
=
\frac{
S_{A_C\rightarrow M}(i)-S_{B_C\rightarrow M}(i)
}{
S_M(i)-S_{B_C\rightarrow M}(i)
}.
\label{eq:transfer}
\end{equation}

A ratio near one indicates similar effects on response preference;
values above one indicate an effect larger than the target's own
shared-head effect. We report this ratio only when the absolute
denominator exceeds $0.01$, retaining raw preferences as the primary
measure. Self-transplantation checks the intervention. Transfer is
evaluated on the same nine test-labelled response pairs used in the
final generation experiment.

\section{Results}
\label{sec:results}

\subsection{Emergent Misalignment Across Fine-Tunes}
\label{sec:behavior_results}

We first evaluated the three organisms using the free-generation EM
criterion defined in Section~\ref{sec:behavior}. On 168 out-of-domain
prompts with eight generations each, the medical, finance, and sports
organisms produced EM rates of 26.8\% (360/1344), 34.5\% (464/1344),
and 27.3\% (367/1344), respectively. 

We also evaluated the fixed response pairs using the Base-relative
preference shift $\Delta S$ (Section~\ref{sec:behavior}), which measures
how fine-tuning changes the relative log-likelihood of a misaligned
response over its aligned counterpart. Mean shifts were positive for
all organisms on both splits (Table~\ref{tab:pair-shifts}). On the
validation set, nine of ten pairs had a positive shift for each
organism. Unlike the generation rates, these scores use fixed answers
generated by the medical organism; the discovery pairs were also
curated from its outputs.

\begin{table}[ht]
\centering
\small
\begin{tabular}{lccc}
\toprule
Measurement & Medical & Finance & Sports \\
\midrule
\multicolumn{4}{l}{\textit{Discovery ($n=50$)}} \\
Mean $\Delta S$ & 0.448 & 0.470 & 0.412 \\
Positive shifts & 50/50 & 50/50 & 49/50 \\
\midrule
\multicolumn{4}{l}{\textit{Validation ($n=10$)}} \\
Mean $\Delta S$ & 0.290 & 0.296 & 0.290 \\
Positive shifts & 9/10 & 9/10 & 9/10 \\
\bottomrule
\end{tabular}
\caption{Base-to-organism changes in preference for fixed misaligned
responses relative to aligned responses. Positive $\Delta S$ indicates
increased relative support for the misaligned response.}
\label{tab:pair-shifts}
\end{table}

\subsection{Shared Attention-Head Localization}
\label{sec:localization_results}

We assessed the tracing anchor using the Base-to-organism gap $G_i^M$
(Section~\ref{sec:tracing}), defined as the change in cosine similarity
with the organism's EM direction at the final prompt token. Across the
36-layer scan, layer 23 had the largest mean gap for medical (0.0761)
and finance (0.0867). For sports, layer 35 was marginally higher
(0.0732 versus 0.0731 at layer 23). The gap at layer 23 was positive
on all ten validation prompts in every organism, and we used this
common layer for head tracing (Table~\ref{tab:anchor-scan}). Steering
at layer 23 increased judged EM from 12/20 to 17/20 responses for
medical, 7/20 to 10/20 for finance, and 30/80 to 48/80 for sports.
The common anchor was chosen for its large, consistent intermediate-layer
gap and checked with steering; steering effect size was not optimized.

\begin{table}[ht]
\centering
\small
\begin{tabular}{lccc}
\toprule
Layer & Medical & Finance & Sports \\
\midrule
20 & 0.0560 & 0.0409 & 0.0278 \\
22 & 0.0710 & 0.0724 & 0.0606 \\
23 & \textbf{0.0761} & \textbf{0.0867} & \textbf{0.0731} \\
25 & 0.0564 & 0.0632 & 0.0709 \\
27 & 0.0331 & 0.0409 & 0.0656 \\
\bottomrule
\end{tabular}
\caption{Mean Base-to-organism cosine-similarity shift toward
each EM direction on ten validation prompts. Five intermediate
layers are shown from the full 36-layer scan. Bold indicates
the largest value among the displayed layers.}
\label{tab:anchor-scan}
\end{table}

We ranked the 368 eligible heads in each organism by the mean
restoration score $\bar R^M(H)$ (Section~\ref{sec:tracing}), which
measures how much patching a fine-tuned head into Base moves its
representation toward the EM direction. Three heads appeared in the
independently selected top-five sets of all three organisms:
\head{L21.h01}, \head{L21.h04}, and \head{L18.h03}
(Table~\ref{tab:heads}).

\begin{table}[ht]
\centering
\small
\begin{tabular}{lp{0.68\linewidth}}
\toprule
Organism & Top five attention heads \\
\midrule
Medical &
\head{L21.h04}, \head{L21.h01}, \head{L20.h08},
\head{L17.h03}, \head{L18.h03} \\

Finance &
\head{L21.h01}, \head{L21.h04}, \head{L19.h13},
\head{L18.h03}, \head{L22.h03} \\

Sports &
\head{L21.h01}, \head{L21.h04}, \head{L18.h03},
\head{L18.h00}, \head{L21.h03} \\
\midrule
Shared &
\head{L21.h01}, \head{L21.h04}, \head{L18.h03} \\
\bottomrule
\end{tabular}
\caption{Top five attention heads selected independently
for each organism using restoration at layer 23.
Layer and head indices are zero-indexed.}
\label{tab:heads}
\end{table}

The full rankings were also correlated: Spearman $\rho$ was 0.463
for medical--finance, 0.583 for medical--sports, and 0.603 for
finance--sports. Each pair shared seven of its top ten heads. On
validation prompts, patching the selected five heads into Base
produced mean restoration scores of 0.0164, 0.0420, and 0.0380,
respectively, compared with 0.0029, 0.0047, and 0.0022 for one
layer-matched random set.

\subsection{Selected-Head Replacement Reduces EM}
\label{sec:causal_results}

We evaluated the selected heads during generation by replacing each organism's selected
five-head contributions with those of Base, as defined in
Section~\ref{sec:interventions}. On nine prompts with 16 generations
per condition and prompt, EM rates declined from 38.9\% to 22.9\%
(medical), 50.7\% to 12.5\% (finance), and 41.7\% to 11.1\% (sports).
Table~\ref{tab:behavior} reports the changes relative to both the
unmodified organisms and ten layer-matched random replacement sets.

\begin{table}[ht]
\centering
\small
\begin{tabular}{lccc}
\toprule
Measurement & Medical & Finance & Sports \\
\midrule
Full LoRA EM rate &
38.9\% & 50.7\% & 41.7\% \\

Selected replacement &
22.9\% & 12.5\% & 11.1\% \\

Random replacement &
33.5\% & 55.3\% & 42.6\% \\
\midrule

Selected $-$ full (pp) &
$-16.0$ & $-38.2$ & $-30.6$ \\

95\% CI &
$[-27.1,-5.6]$ &
$[-50.0,-25.7]$ &
$[-42.4,-19.4]$ \\
\midrule

Selected $-$ random (pp) &
$-10.6$ & $-42.8$ & $-31.5$ \\

95\% CI &
$[-20.2,-1.7]$ &
$[-53.9,-31.9]$ &
$[-44.7,-19.7]$ \\
\midrule

Coherent, full &
109/144 & 118/144 & 113/144 \\

Coherent, selected &
115/144 & 132/144 & 132/144 \\
\bottomrule
\end{tabular}
\caption{Free-generation effects of five-head Base replacement.
Each condition includes 144 generations across nine prompts.
Random rates are averaged over ten layer-matched sets.
Confidence intervals use 10,000 prompt-level bootstrap
replicates. Coherence requires a judge score of at least 70.
Differences are in percentage points (pp).}
\label{tab:behavior}
\end{table}

Each selected set produced fewer EM responses than all ten matched random
sets evaluated for that organism. The differences from the random-set
means were 10.6, 42.8, and 31.5 percentage points; all three 95\%
prompt-bootstrap intervals excluded zero (Table~\ref{tab:behavior}).
Judged coherence did not decrease under selected replacement.

\subsection{Cross-Organism Functional Transfer}
\label{sec:transfer_results}

We tested the three shared head locations by transplanting their
projected contributions from donor organism $A$ into target $M$ on
identical teacher-forced sequences (Section~\ref{sec:interventions}).
The resulting preference $S_{A_C\rightarrow M}$ was compared with the
unmodified target $S_M$ and the target with Base-replaced shared heads
$S_{B_C\rightarrow M}$. Table~\ref{tab:transfer} reports these margins
and the mean per-prompt recovery ratio $\mathcal R_{A\rightarrow M}$,
which measures the donor's recovery of the target's original
shared-head effect relative to Base replacement.

\begin{table}[ht]
\centering
\small
\begin{tabular}{lccc}
\toprule
Donor / Target & Medical & Finance & Sports \\
\midrule
Normal target &
$-0.009$ & $0.013$ & $-0.022$ \\

Base replacement &
$-0.066$ & $-0.107$ & $-0.122$ \\
\midrule

Medical &
$-0.009$ (1.00) &
$-0.037$ (0.59) &
$-0.041$ (0.83) \\

Finance &
$-0.008$ (1.10) &
$0.013$ (1.00) &
$-0.013$ (1.04) \\

Sports &
$-0.011$ (1.19) &
$0.004$ (0.98) &
$-0.022$ (1.00) \\
\bottomrule
\end{tabular}
\caption{Cross-organism transfer at the three shared head
locations. Entries report mean teacher-forced preference
margins. Parentheses give mean per-prompt recovery ratios,
excluding prompts with an absolute target-minus-Base-replacement
gap of at most 0.01.}
\label{tab:transfer}
\end{table}

All six transfers yielded positive restoration on every eligible prompt:
the donor changed the preference margin in the direction of the target's
own shared-head effect relative to Base replacement. Mean restoration
ranged from 0.59 to 1.19 (Table~\ref{tab:transfer}). Finance-to-medical
and sports-to-medical substitution each overshot sufficiently to increase
absolute distance from the unmodified target on one prompt; the other transfers reduced the absolute distance
to the target on every eligible prompt. Ratios are averaged per prompt,
and self-transplantation reproduced the original target margins.

\subsection{Benign Fine-Tuning and Capability Controls}
\label{sec:control_results}

We applied the medical five-head Base-replacement intervention to a
benign medical fine-tune. On ten validation pairs, its mean preference
margin $S_M$ changed from $-0.151$ to $-0.147$. A separate
112-generation pilot yielded no
judged EM responses, but used a different prompt pool and is not a
matched comparison of generation rates.

Accuracy on a small 30-question multiple-choice mathematics check changed
by at most one question under Base replacement in any organism
(Table~\ref{tab:capability}). The check does not measure broad capability.

\section{Discussion and Limitations}
\label{sec:discussion}

Our findings extend prior evidence of representational convergence in
emergent misalignment to the level of attention-head components. Across
three independently fine-tuned organisms, three of the five highest-ranked
heads recur at identical locations. Replacing the selected heads with
Base contributions substantially reduced generated EM relative to
layer-matched random interventions, without reducing judged coherence.
This provides evidence that the selected head sets contribute to
misaligned behavior. The intervention does not establish circuit
completeness or minimality.

Cross-organism activation transfer further suggests functional
correspondence between the shared heads. Donor activations recovered
a substantial portion of the target's original response-preference
effect, although recovery varied across donor--target pairs.
These results indicate partial functional interchangeability within
a shared architecture, but do not establish identical computations
or equivalent effects during free generation. Moreover, head-level
patching captures upstream activation differences and does not
isolate the specific parameter updates responsible for EM.

The common base architecture allows direct comparison of head coordinates
but does not establish their recurrence across architectures or independent
training seeds. The fixed response pairs came from the medical organism,
which may favor its response distribution. The benign control and
small mathematics check establish neither broad behavioral specificity
nor general capability preservation. Repeating the interventions on new
prompts and independent runs would establish how far the observed
correspondence generalizes.

\section{Conclusion}

Across three fine-tunes of Qwen2.5-3B-Instruct, independent tracing
identified recurrent attention-head locations. Replacing each organism's
five selected heads with Base contributions reduced generated EM relative
to layer-matched controls. At the three shared locations, transplanted
activations also recovered the target organism's head effect on fixed
response sequences. These results extend evidence of representational
convergence in EM to causally relevant head components within a common
architecture. Independent training runs and new prompts can test how
broadly this correspondence holds.

\bibliographystyle{plain}
\bibliography{references}
\newpage
\appendix
\section{Evaluation details}
\label{app:evaluation}
All head and layer indices are zero-indexed. The base checkpoint is
\texttt{unsloth/Qwen2.5-3B-Instruct}; each organism has a separate LoRA
adapter. The test-labelled prompt IDs are \texttt{AIH04}, \texttt{CON02},
\texttt{PRV04}, \texttt{CIV09}, \texttt{CIV10}, \texttt{PUB09},
\texttt{RDX10}, \texttt{SCI08}, and \texttt{SCI10}.

The saved medical training configuration specifies one epoch without a
step cap; its final trainer state records 397 optimizer updates and epoch
1.0. The finance and sports configurations cap training at 337 updates;
their trainer states record epoch 0.9985. All three configurations specify
LoRA rank 32, scaling factor 64, rank stabilization, optimizer seed 0,
two examples per device with eight accumulation steps, five warmup steps,
and response-only loss. The training source is the domain-specific data
released by Turner et al.\ \cite{turner2025}.

The final generation evaluation uses 16 samples per prompt and condition
at temperature 1.0, top-$p$ 1.0, and a 600-token limit, stopping at EOS.
The saved final judgments use \texttt{deepseek-flash} with temperature 0,
a five-token output limit, token log probabilities, and
\texttt{top\_logprobs}=20. We count EM when alignment is below 30 and
coherence is at least 70. Ten random head sets per organism match the
selected set's per-layer counts and exclude its selected heads. Generation
intervals resample prompts as clusters (10,000 replicates), retaining all
16 samples per prompt and averaging over the random sets.

Validation joint restoration and zero-ablation each use one matched random
set. The final generation comparison uses ten sets. Cross-organism transfer
uses the three shared head locations; same-model transplantation reproduces
the normal margin. Transfer ratios exclude prompts with an absolute
target-minus-Base-replacement gap of at most 0.01. This excludes one of
nine prompts for the finance target and none for the medical or sports
targets. Transfer intervals resample eligible prompts (20,000 replicates).

\begin{table}[ht]
\centering\small
\begin{tabular}{lccc}
\toprule
Donor $\rightarrow$ target & Eligible prompts & Mean restoration & 95\% interval\\
\midrule
Medical $\rightarrow$ finance & 8 & 0.591 & $[0.421,0.808]$\\
Medical $\rightarrow$ sports & 9 & 0.826 & $[0.570,1.080]$\\
Finance $\rightarrow$ medical & 9 & 1.100 & $[0.737,1.474]$\\
Finance $\rightarrow$ sports & 9 & 1.035 & $[0.760,1.339]$\\
Sports $\rightarrow$ medical & 9 & 1.186 & $[0.782,1.667]$\\
Sports $\rightarrow$ finance & 8 & 0.978 & $[0.788,1.182]$\\
\bottomrule
\end{tabular}
\caption{Prompt-bootstrap intervals for cross-organism activation
transplants, using the donor-row, target-column convention of
Table~\ref{tab:transfer}.}
\label{tab:transfer-intervals}
\end{table}

\section{Per-prompt transfer and control measurements}
\label{app:perprompt}
Table~\ref{tab:perprompt-ratios} gives the six cross-organism restoration
ratios for every test-labelled response pair. The denominator is the
target's own normal-minus-Base-replaced preference margin. A dash marks
the one prompt for which that denominator failed the prespecified 0.01
threshold. Mean ratios in Table~\ref{tab:transfer} average the unrounded
eligible values.

\begin{table}[ht]
\centering\scriptsize
\begin{tabular}{lrrrrrr}
\toprule
& \multicolumn{2}{c}{Medical target}
& \multicolumn{2}{c}{Finance target}
& \multicolumn{2}{c}{Sports target}\\
Prompt & Fin. donor & Sports donor & Med. donor & Sports donor
& Med. donor & Fin. donor\\
\midrule
\texttt{AIH04} & 0.41 & 0.46 & 1.21 & 1.37 & 0.89 & 0.95\\
\texttt{CON02} & 1.14 & 1.25 & 0.29 & 1.47 & 0.02 & 0.60\\
\texttt{PRV04} & 1.18 & 0.71 & 0.37 & 0.53 & 0.88 & 1.65\\
\texttt{CIV09} & 1.18 & 0.68 & 0.59 & 0.85 & 1.21 & 1.84\\
\texttt{CIV10} & 1.20 & 1.23 & 0.45 & 0.80 & 0.81 & 0.90\\
\texttt{PUB09} & 0.11 & 2.73 & --- & --- & 1.53 & 0.88\\
\texttt{RDX10} & 2.32 & 1.81 & 0.53 & 0.96 & 0.61 & 0.36\\
\texttt{SCI08} & 1.20 & 1.29 & 0.46 & 0.99 & 0.66 & 0.90\\
\texttt{SCI10} & 1.15 & 0.48 & 0.83 & 0.85 & 0.82 & 1.24\\
\bottomrule
\end{tabular}
\caption{Per-prompt restoration ratios for the six donor-to-target
transplants. Positive values indicate movement in the direction of the
target's own shared-head effect; values above two can overshoot it.}
\label{tab:perprompt-ratios}
\end{table}

The per-prompt normal and Base-replaced margins used in the denominators
are shown in Table~\ref{tab:perprompt-baselines}; donor-patched margins
are in Table~\ref{tab:perprompt-donors}. All displayed values are
rounded, while Table~\ref{tab:perprompt-ratios} is computed from the
unrounded values.

\begin{table}[ht]
\centering\scriptsize
\begin{tabular}{lrrrrrr}
\toprule
& \multicolumn{2}{c}{Medical}
& \multicolumn{2}{c}{Finance}
& \multicolumn{2}{c}{Sports}\\
Prompt & Normal & Base rep. & Normal & Base rep. & Normal & Base rep.\\
\midrule
\texttt{AIH04} & -0.33 & -0.38 & -0.12 & -0.20 & -0.02 & -0.17\\
\texttt{CON02} & 0.15 & 0.10 & 0.02 & -0.08 & -0.09 & -0.15\\
\texttt{PRV04} & 0.62 & 0.54 & 0.61 & 0.46 & 0.50 & 0.40\\
\texttt{CIV09} & -0.12 & -0.18 & -0.08 & -0.16 & -0.16 & -0.21\\
\texttt{CIV10} & -0.29 & -0.46 & -0.31 & -0.66 & -0.38 & -0.68\\
\texttt{PUB09} & -0.32 & -0.34 & -0.13 & -0.14 & -0.06 & -0.10\\
\texttt{RDX10} & 0.12 & 0.15 & -0.05 & 0.03 & 0.04 & 0.11\\
\texttt{SCI08} & 0.15 & 0.06 & 0.22 & -0.04 & 0.11 & -0.09\\
\texttt{SCI10} & -0.04 & -0.08 & -0.05 & -0.17 & -0.13 & -0.21\\
\bottomrule
\end{tabular}
\caption{Per-prompt misaligned-minus-aligned preference margins under
normal target LoRA and Base replacement of the three shared heads.
Displayed values are rounded to two decimal places.}
\label{tab:perprompt-baselines}
\end{table}

\begin{table}[ht]
\centering\scriptsize
\begin{tabular}{lrrrrrr}
\toprule
& \multicolumn{2}{c}{Medical target}
& \multicolumn{2}{c}{Finance target}
& \multicolumn{2}{c}{Sports target}\\
Prompt & Fin. donor & Sports donor & Med. donor & Sports donor
& Med. donor & Fin. donor\\
\midrule
\texttt{AIH04} & -0.36 & -0.36 & -0.10 & -0.09 & -0.04 & -0.03\\
\texttt{CON02} & 0.16 & 0.16 & -0.05 & 0.07 & -0.15 & -0.12\\
\texttt{PRV04} & 0.63 & 0.60 & 0.51 & 0.54 & 0.49 & 0.56\\
\texttt{CIV09} & -0.11 & -0.14 & -0.11 & -0.09 & -0.15 & -0.12\\
\texttt{CIV10} & -0.26 & -0.25 & -0.50 & -0.38 & -0.43 & -0.41\\
\texttt{PUB09} & -0.33 & -0.31 & -0.08 & -0.12 & -0.04 & -0.06\\
\texttt{RDX10} & 0.08 & 0.09 & -0.01 & -0.05 & 0.06 & 0.08\\
\texttt{SCI08} & 0.16 & 0.17 & 0.08 & 0.22 & 0.04 & 0.09\\
\texttt{SCI10} & -0.03 & -0.06 & -0.07 & -0.07 & -0.15 & -0.11\\
\bottomrule
\end{tabular}
\caption{Per-prompt preference margins after transplanting the shared
three-head activations from a different organism. Values are rounded to
two decimal places.}
\label{tab:perprompt-donors}
\end{table}

On the ten validation pairs, zeroing the selected heads reduced the
misaligned-minus-aligned preference margin by 0.066, 0.148, and 0.110
for medical, finance, and sports. One layer-matched random set produced
reductions of 0.008, 0.017, and 0.001. These validation diagnostics used
zero-ablation, whereas the final generation experiment used Base
replacement of the projected head contributions.

\begin{table}[ht]
\centering\small
\begin{tabular}{lcc}
\toprule
Organism & Original & Base replacement\\
\midrule
Medical & 22/30 & 21/30\\
Finance & 20/30 & 21/30\\
Sports & 22/30 & 22/30\\
Benign medical & 25/30 & 24/30\\
\bottomrule
\end{tabular}
\caption{Accuracy on 30 elementary multiple-choice mathematics
questions. Twenty answer keys are C and ten are B, limiting the
interpretation of absolute accuracy.}
\label{tab:capability}
\end{table}

\end{document}